\documentclass[10pt,twocolumn,letterpaper]{article}

\usepackage[pagenumbers]{templates/ICCV2025-Author-Kit-Feb/iccv}      

\usepackage{siunitx}
\usepackage{float}
\usepackage{dblfloatfix}
\usepackage{colortbl}
\PassOptionsToPackage{usenames,dvipsnames,table}{xcolor}
\usepackage{appendix}
\usepackage[accsupp]{axessibility}
\usepackage{multirow}

\definecolor{iccvblue}{rgb}{0.21,0.49,0.74}
\usepackage[pagebackref,breaklinks,colorlinks,allcolors=iccvblue]{hyperref}

\title{EvRT-DETR: Latent Space Adaptation of Image Detectors\\ for Event-based Vision}

\author{%
{Dmitrii Torbunov}$^1$\thanks{© 2025 IEEE},
{Yihui Ren}$^1$,
{Animesh Ghose}$^1$,
{Odera Dim}$^1$,
{Yonggang Cui}$^1$\\[1ex]
$^1$Brookhaven National Laboratory, Upton, NY, USA\\[1ex]
{\tt\small dtorbunov@bnl.gov, yren@bnl.gov, aghose@bnl.gov, dodera@bnl.gov, ycui@bnl.gov}
}

\newcommand{\thename}[0]{EvRT-DETR\xspace}
\newcommand{\therepo}[0]{\url{https://github.com/realtime-intelligence/evrt-detr}}

\newcommand{\tempname}[0]{I2EvDet\xspace}

\newcommand{\genOne}[0]{\mbox{Gen1}\xspace}
\newcommand{\genFour}[0]{\mbox{1Mpx}\xspace}

\begin{document}

\maketitle

\begin{abstract}

Event-based cameras (EBCs) have emerged as a bio-inspired alternative to traditional cameras, offering advantages in power efficiency, temporal resolution, and high dynamic range.
However, development of image analysis methods for EBCs is challenging due to the sparse and asynchronous nature of the data.
This work addresses the problem of object detection for EBC cameras.
The current approaches to EBC object detection focus on constructing complex data representations and rely on specialized architectures.
We introduce \tempname (Image-to-Event Detection), a novel adaptation framework that bridges mainstream object detection with temporal event data processing.
First, we demonstrate that a Real-Time DEtection TRansformer, or RT-DETR, a state-of-the-art natural image detector, trained on a simple image-like representation of the EBC data achieves performance comparable to specialized EBC methods.
Next, as part of our framework, we develop an efficient adaptation technique that transforms image-based detectors into event-based detection models by modifying their frozen latent representation space via minimal architectural additions.
The resulting \thename model reaches state-of-the-art performance on the standard benchmark datasets \genOne (mAP $+2.3$) and \mbox{1Mpx/Gen4} (mAP $+1.4$).
These results demonstrate a fundamentally new approach to EBC object detection through principled adaptation of mainstream architectures, offering an efficient alternative with potential applications to other temporal visual domains.
The code is available at: \therepo.

\end{abstract}
\section{Introduction}

\begin{figure}[hth!]
    \centering
    \includegraphics[width=1.0\columnwidth]{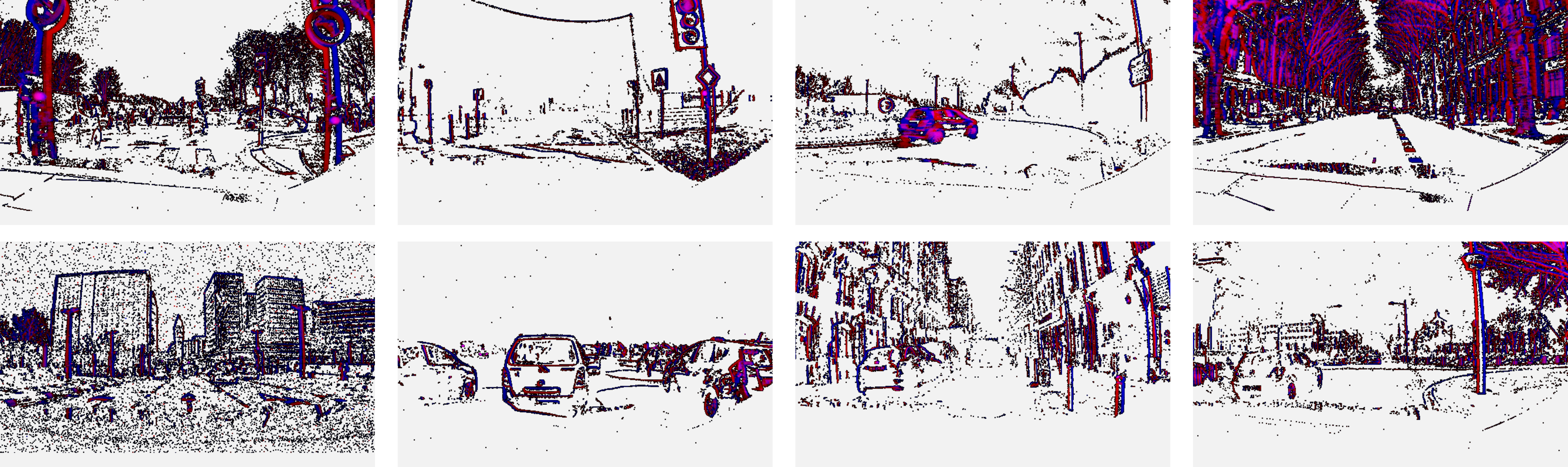}
    \caption{
        \textbf{Event-based Camera Output Visualization.} Sample frames from the \genFour dataset showing event polarity. Red indicates positive events, and blue denotes negative events.}
    \label{fig:image_grid}
\end{figure}

Event-based cameras (EBCs) present a biologically inspired alternative to traditional frame-based cameras. Unlike conventional cameras that capture frames at predetermined intervals, EBC pixels operate asynchronously, generating data only when brightness changes exceed a threshold. This results in sparse, asynchronous data streams where each event is a tuple containing pixel location, timestamp, and polarity (whether brightness increased or decreased), as illustrated in \autoref{fig:image_grid}. This approach offers remarkable advantages: low power consumption (as low as \si{10} {mW}), reduced data transfer rates, exceptional temporal resolution (on the order of $\mu s$), and high dynamic range ($>$\si{100} {dB}). These properties have led to widespread adoption in autonomous driving, robotics, and wearable electronics~\cite{gallego2020event}.

\begin{figure}[t]
    \centering
    \includegraphics[width=1.0\columnwidth]{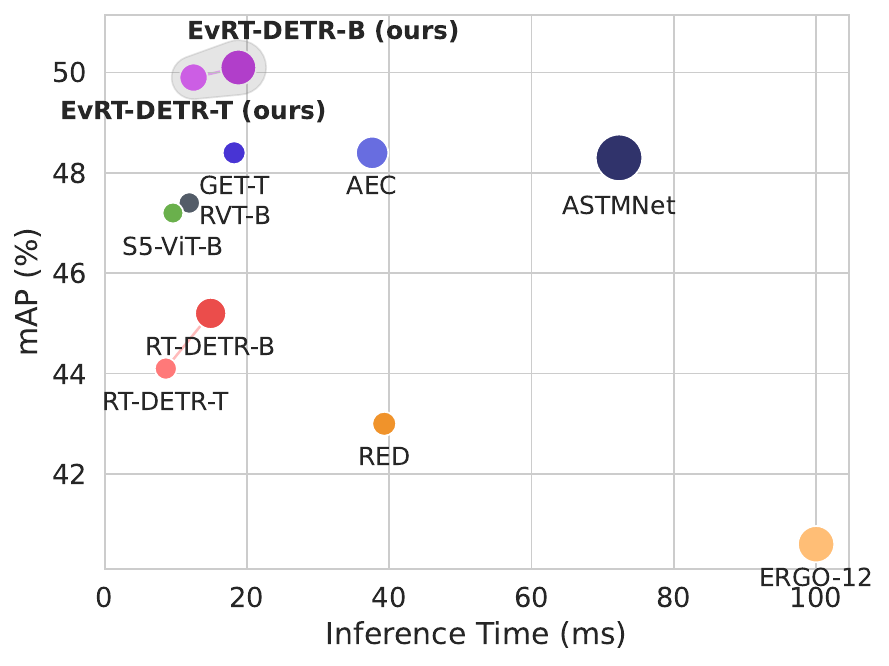}
    \caption{
        \textbf{Object Detection Performance vs. Inference Time.}
        A summary of object detection performance (COCO mAP) versus inference time (ms) of various models on the \genFour automotive dataset. Circle size is proportional to the number of model parameters. Inference times are reported for NVIDIA T4 GPU.
        Our \thename models achieve state-of-the-art accuracy while maintaining competitive inference speeds.}
    \label{fig:map_vs_time}
\end{figure}

These unique characteristics of EBC data – sparsity, asynchronicity, and temporal nature – create fundamental challenges when applying traditional computer vision techniques. Early attempts to directly apply conventional image-based object detectors to simple event representations yielded poor performance~\cite{perot2020learning}, highlighting the gap between these domains. This led to two predominant research directions: (1) constructing sophisticated event representations that transform EBC data into formats compatible with existing detectors (e.g.,~\cite{zubic2023chaos}) or (2) designing specialized architectures tailored to handle the temporal nature of event data (e.g.,~\cite{peng2023get}). A common assumption in these approaches is that EBC data processing requires either specialized representations or architectures due to the fundamental differences between event- and frame-based vision, leading to development paths that often diverge from mainstream computer vision research.

Meanwhile, mainstream object detection has seen remarkable progress, evolving from convolutional neural network (CNN)-based architectures, e.g., YOLO~\cite{redmon2016you} and Faster R-CNN~\cite{ren2016faster}, to transformer-based models like DETR (DEtection TRansformer)~\cite{carion2020end}. Recent advances have culminated in highly efficient and accurate models, such as the \textbf{R}eal-\textbf{T}ime \textbf{DE}tection \textbf{TR}ansformer (RT-DETR)~\cite{zhao2024detrs} that achieves exceptional performance while maintaining real-time inference speeds. Though primarily optimized for processing individual images, these state-of-the-art detectors offer powerful feature extraction capabilities that potentially could be leveraged for temporal data.

This evolution in mainstream detection creates an opportunity to bridge the gap between conventional and event-based vision. Rather than assuming fundamental incompatibility between these domains, we explore a complementary direction: adapting state-of-the-art frame-based detectors to process event data effectively with minimal modifications. In this work, we investigate how RT-DETR can be strategically adapted to the event-based domain, demonstrating that modern object detection architectures can successfully work on event data and opening a new research direction.

Our approach proceeds in two stages to systematically adapt RT-DETR to event data.
First, we train RT-DETR directly on simple image-like representations of EBC
data using standard training procedures without special modifications. This
straightforward approach yields surprising results -- even when processing only
fixed \SI{50}{ms} time frames, the detector achieves performance comparable to
specialized EBC-specific methods explicitly designed for video object
detection on EBC data. This is particularly unexpected as these specialized
methods deliberately exploit the temporal nature of EBC data and process
information from the distant past, while our approach performs detection using
only a single fixed time window (frame) of $\SI{50}{ms}$, similar to
traditional frame-based cameras.

Building on this foundation, we introduce \textbf{\tempname (Image-to-Event Detection)}, a technique that enables straightforward transformation of image-based object detectors into video-capable models. This approach leverages latent space adaptation principles~\cite{houlsby2019parameter,hu2021lora} where we freeze the pre-trained RT-DETR model and insert a lightweight recurrent neural network (RNN) module within the encoder's latent representation space.
This design allows the model
to capture temporal dynamics across frames while preserving the powerful
representation capabilities of the original detector and adding minimal
computational overhead. The resulting adaptation effectively leverages
information across multiple time windows without requiring a specialized video
detection architecture.

The combined approach, called \textbf{\thename}, achieves state-of-the-art performance on standard EBC benchmarks, outperforming specialized methods on both \genOne~\cite{de2020large} (mAP $+2.3$) and \genFour~\cite{perot2020learning} (mAP $+1.4$) datasets. Importantly, the results are achieved using only standard modules from natural image and video analysis \textit{without} requiring domain-specific modifications. This demonstrates not only the effectiveness of our approach for EBC data, but also suggests a broader paradigm for adapting image detectors to temporal domains. As shown in Figure~\ref{fig:map_vs_time}, our model achieves superior accuracy while maintaining competitive inference speeds, making it practical for real-world applications.

Contributions of this work:
\begin{itemize}
    \item We show that when trained on simple image-like representations of EBC data using standard procedures, mainstream object detectors like RT-DETR can achieve performance comparable to specialized EBC methods, offering a complementary approach to domain-specific architectures.
    
    \item We propose \tempname, an efficient adaptation technique that transforms image-based detectors into video-capable models through minimal architectural additions to the frozen pre-trained model and with potential applications to various temporal visual domains.
    
    \item We validate our approach in the challenging domain of event-based cameras, where \thename achieves state-of-the-art performance on the standard \genOne (mAP $+2.3$) and \genFour (mAP $+1.4$) EBC benchmarks while maintaining competitive inference speeds.
\end{itemize}

\section{Related Work}

Research at the intersection of event-based cameras and object detection spans multiple domains in computer vision. Current approaches primarily focus on two directions: designing efficient representations for the unique properties of event data and developing specialized neural architectures to process these representations. However, despite its success in other domains, a third direction -- adapting mainstream computer vision models to event data -- remains underexplored. This section reviews relevant work across these areas, highlighting how our approach bridges the gap between specialized event-based methods and mainstream object detection advances.

\subsection{EBC Data Representations}

An event camera produces a stream of event data in the form $(t, p, x, y)$, where $t$ is the timestamp of an event, $p$ is the polarity (positive or negative) of the brightness change, and $(x, y)$ is the location of the event pixel.
To simplify the analysis, the event data often are transformed into alternative representations more suitable for conventional image analysis algorithms.

The simplest image-like representations of event camera data are \textit{Event Frames} or \textit{2D Histograms}~\cite{gallego2020event}, where events within fixed time windows are accumulated into two-dimensional frames of shape $(2, H, W)$ with polarity as the first dimension and spatial coordinates as the remaining dimensions. A natural evolution is the \textit{Stacked 2D Histogram}, which further partitions each frame into $T$ time intervals, creating a representation of shape $(2, T, H, W)$ that typically is reshaped to $(2T, H, W)$ and treated as a natural image. While these representations are straightforward, directly applying existing computer vision algorithms to them has yielded poor performance in the past~\cite{perot2020learning}.

To improve performance, other representations are being explored.
For example, a \textit{Time Surface (TS)} representation~\cite{zubic2023chaos,gallego2020event} is an image-like representation where pixel values encode the time elapsed since the last event in a given pixel.
Thus, unlike the simplest fixed time window representations, \textit{TS} potentially can encode arbitrarily distant past.
Empirically, it has been shown that using the \textit{TS} representations offers significantly better performance on the object detection tasks compared to the \textit{2D Histogram} representations~\cite{perot2020learning}.

Many other EBC data representations also have been proposed.
The recent ERGO-12 (Event Representation through Gromov-Wasserstein Optimization)~\cite{zubic2023chaos} work has developed efficient image-like representations, achieving state-of-the-art object detection performance on the \genOne dataset.
Meanwhile, other approaches reformulate object detection as a three-dimensional (3D) problem by treating events as points in spatial-temporal space or discretizing them into voxel grids~\cite{gallego2020event}.
While computationally heavier, these representations can preserve exact event timing, potentially enabling better handling of overlapping objects and complex motions.

\subsection{EBC Object Detection Architectures}

Once the data representation is selected, one can start to develop object detection algorithms.
There are several families of object detection methods designed for particular data representations.

The most straightforward approach applies conventional image detection methods directly to \textit{2D Histogram} frames, but this yields relatively poor performance~\cite{perot2020learning}. More advanced methods, such as Recurrent Vision Transformer (RVT)~\cite{gehrig2023recurrent} and State Space Vision Transformer (S5-ViT-B)~\cite{zubic2024state}, extend this approach by incorporating temporal memory to capture sequential information, significantly improving detection quality compared to frame-by-frame processing.

The most recent family of methods attempts to use more efficient event representations to further improve object detection performance.
For example, 12-channel ERGO-12~\cite{zubic2023chaos} constructs an optimized image-like event representation and shows that the standard object detection methods perform exceptionally well on it.
Alternatively, Asynchronous Spatio-Temporal Memory Network (ASTMNet)~\cite{li2022asynchronous} and Adaptive Event Conversion (AEC)~\cite{peng2023better} attempt to combine better event representation while using novel neural architectures to expand state-of-the-art results.

Other methods have been developed that treat EBC object detection as a 3D detection problem~\cite{schaefer2022aegnn,perot2020learning,gehrig2022pushing}, exploit benefits of neuromorphic architectures~\cite{cordone2022object}, or explore various hybrid approaches~\cite{peng2023get,peng2024scene}.

\begin{figure*}[htb]
    \centering
    \includegraphics[width=1.0\textwidth]{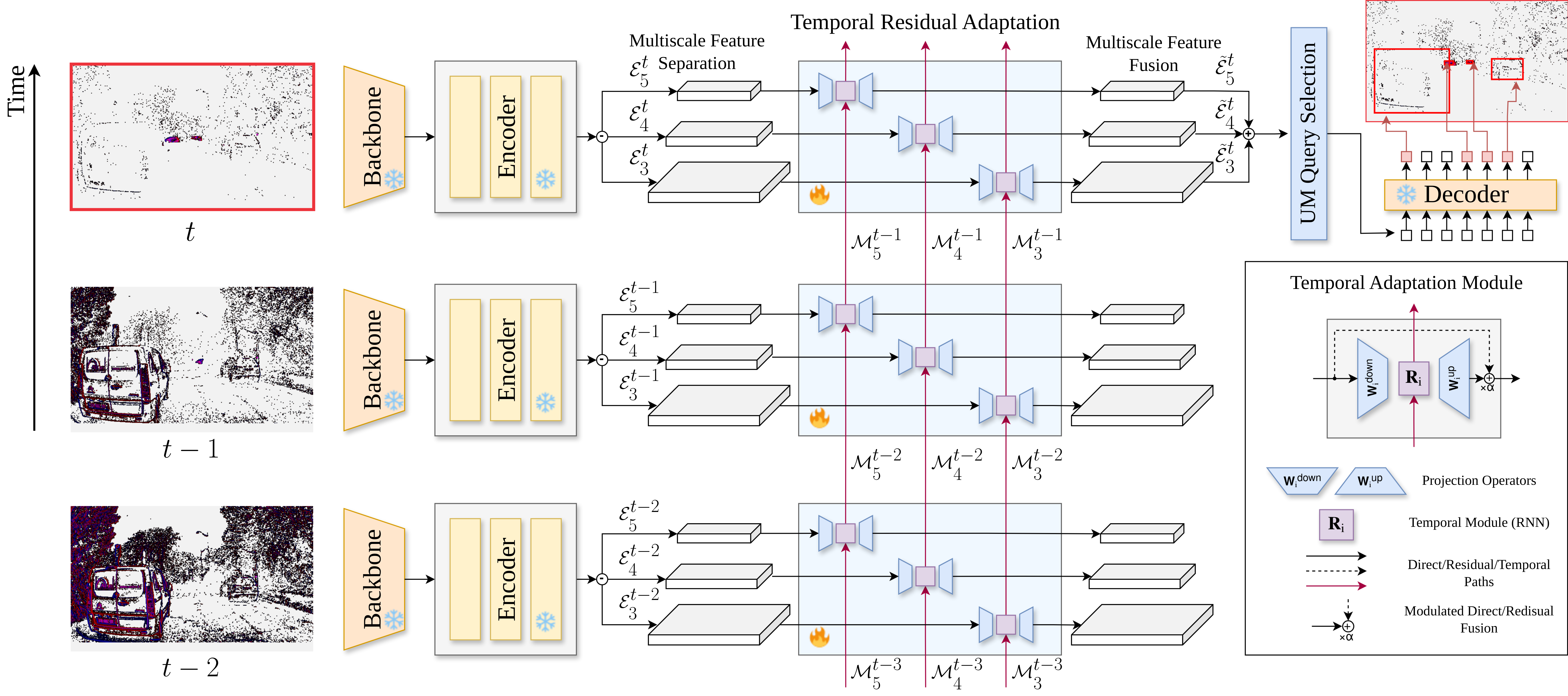}
    \caption{
    \textbf{Overview of the \tempname Framework Applied to RT-DETR}.
    This approach adapts a pre-trained RT-DETR model to process temporal event data using minimal architectural modifications.
    The backbone and encoder components remain frozen during temporal adaptation, while trainable temporal modules are strategically inserted at multiple feature scales.
    Left: Event camera frames are processed through the frozen RT-DETR backbone and encoder. Center: Our temporal adaptation modules operate on multiscale features through residual connections (red pathways), enabling information flow across time steps while preserving spatial representation integrity.
    Right: Feature fusion and decoder stages produce final detections. The detailed view of the Temporal Adaptation Module (bottom right) shows the projection operations ($W_i^{down}, W_i^{up}$) and recurrent memory module ($\mathbf{R}_i$) that enable efficient temporal modeling with minimal additional parameters.
    }
    \label{fig:evrt_detr}
\end{figure*}

\subsection{Transformer-based Object Detection}

Object detection has been revolutionized by transformer architectures in recent years. DETR~\cite{carion2020end} was the first widely successful transformer-based~\cite{vaswani2017attention} model for object detection, offering a simple and elegant alternative to traditional CNN-based architectures~\cite{girshick2014rich,redmon2016you}. 

DETR follows a hybrid CNN-Transformer design with a backbone-encoder-decoder structure. It uses a CNN backbone (typically ResNet~\cite{he2016deep}) for feature extraction, followed by a Transformer encoder that processes these features into tokens and captures their correlations. The Transformer decoder then identifies objects through cross-attention to the encoder outputs.

Despite its innovative approach, the original DETR suffered from limitations, including training instability~\cite{zhang2022dino}, poor performance on small objects~\cite{carion2020end}, and high computational costs during inference~\cite{zhu2020deformable}. Subsequent research has addressed these issues through architectural improvements such as Feature Pyramid Networks (FPN), specialized decoder tokens, and deformable attention mechanisms~\cite{zhu2020deformable,carion2020end}.

These incremental advances have culminated in the RT-DETR~\cite{zhao2024detrs}, which achieves exceptional performance across all object scales with stable training and significantly faster inference times. RT-DETR represents the current state-of-the-art in efficient transformer-based detection and serves as an ideal foundation for adaptation to specialized domains, such as event-based vision.

\subsection{Model Adaptation Techniques}

While specialized architectures dominate event-based vision, adapting mainstream architectures offers compelling advantages. Parameter-efficient methods~\cite{houlsby2019parameter,he2021towards,hu2021lora} modify pre-trained models with minimal changes by adding compact modules to frozen backbones. These approaches have proven effective across language~\cite{dettmers2023qlora}, vision~\cite{zhong2024convolution}, and multimodal systems~\cite{han2024parameter}, but they remain underexplored for event-based vision despite their efficiency and transfer learning benefits.

Given the success of adaptation techniques in other domains and the power of transformer-based detection models, there is a clear opportunity to bridge these approaches for event-based vision. Our work explores this intersection, demonstrating that mainstream object detectors can be effectively adapted to EBC data with minimal modifications rather than requiring specialized architectures.

\section{Methods}

This section presents \tempname, our adaptation framework for event-based object detection. Unlike approaches that rely on specialized architectures or complex representations, \tempname bridges mainstream object detection with temporal event data via strategic, minimal modifications.
Our framework demonstrates how image-based detectors can be efficiently transformed into video-capable models while preserving their core strengths. We first describe our image-like representation of event data then introduce the \tempname framework. Finally, we detail the technical implementation and discuss how our approach addresses the unique temporal dependencies in event camera data.

\subsection{Event Representation}
\label{sec:ev_repr}

Our adaptation framework begins with transforming EBC data into an image-like representation that is compatible with mainstream object detection models. We employ the \textit{Stacked 2D Histogram} representation~\cite{gallego2020event} as it provides a straightforward bridge between event data and conventional image processing architectures.

To ensure direct comparability with existing methods, we adopt the frame construction parameters from RVT~\cite{gehrig2023recurrent}, which allows us to isolate the impact of our architectural choices from data pre-processing.
Specifically, we take the stream of events and partition it into a series of consecutive fixed time windows of $T_\text{frame} = \SI{50}{\milli\second}$.
Each such window corresponds to a single frame.
Next, we further subdivide each frame into 10 intervals of $T_\text{bin} = \SI{5}{\milli\second}$.

To construct a frame corresponding to an interval $[ t_0, t_0 + T_\text{frame} ]$, we create an intermediate stacked histogram $S$ from a set of events $\mathcal{E}$ in that interval:
\begin{equation*}
  S(p, t_i, y, x) =
    \sum_{\left\{ e | t_e \in [ t_0, t_0 + T_\text{frame}) \right\}}
        \delta_{x}^{x_e} \delta_{y}^{y_e} \delta_{p}^{p_e}
        \mathbf{1}_{[t_i, t_{i+1})}(t_e)
\end{equation*}
where $e = (t_e, p_e, x_e, y_e)$ represents an individual event with time stamp $t_e$, polarity $p_e$, and spatial coordinates $(x_e, y_e)$; $t_i = t_0 + i \cdot T_\text{bin}$, $i \in [0, 10)$ is the histogram bin index;  $(x, y)$ are the spatial indices in the output histogram; $\delta_{a}^{b}$ is the Kronecker delta function; and $\mathbf{1}$ is an indicator function.
Once a stacked histogram $S$ of shape $(2, 10, H, W)$ is constructed, we merge the polarity and time dimensions to obtain an image-like 2D frame $F$ of shape $(20, H, W)$.

\subsection{\tempname: Image-to-Event Detection Framework}
\label{sec:thename}

\tempname is a principled framework for adapting image-based object detectors to process temporal data. While specialized architectures typically are developed for event-based object detection, this approach demonstrates that strategic adaptation of mainstream detectors can achieve superior performance with minimal architectural modifications.

The \tempname framework applies latent space adaptation principles~\cite{houlsby2019parameter,he2021towards,hu2021lora}, which enable domain adaptation through minimal changes to pre-trained models. The \tempname framework follows a two-stage process, and it is applicable to any object detection model with a clean separation between feature extraction and object detection modules (e.g., YOLO and DETR families).

First, we establish a solid foundation by training an object detector on individual EBC frames (\autoref{sec:ev_repr}), treating them as conventional images. This creates a detector with powerful spatial representation capabilities optimized for event data.

In the second stage, we transform this image-based detector into a video-capable model. We freeze the pre-trained detector parameters and insert lightweight RNN modules between the feature extractor and detection components. We train the inserted temporal modules on the EBC video object detection task while keeping the original detector frozen. The resulting model captures temporal dynamics across frames while preserving the powerful representation capabilities of the original detector and adding minimal computational overhead.

This two-stage approach retains learned spatial representations of the original image detector while efficiently adding temporal capabilities. In this work, we apply \tempname to
the RT-DETR model as the primary detector, resulting in the \thename model.
Additionally, we demonstrate the generalizability of the \tempname framework
using experiments with YOLOX detectors.

\begin{figure*}[t!]
    \centering
    \includegraphics[width=0.9\textwidth]{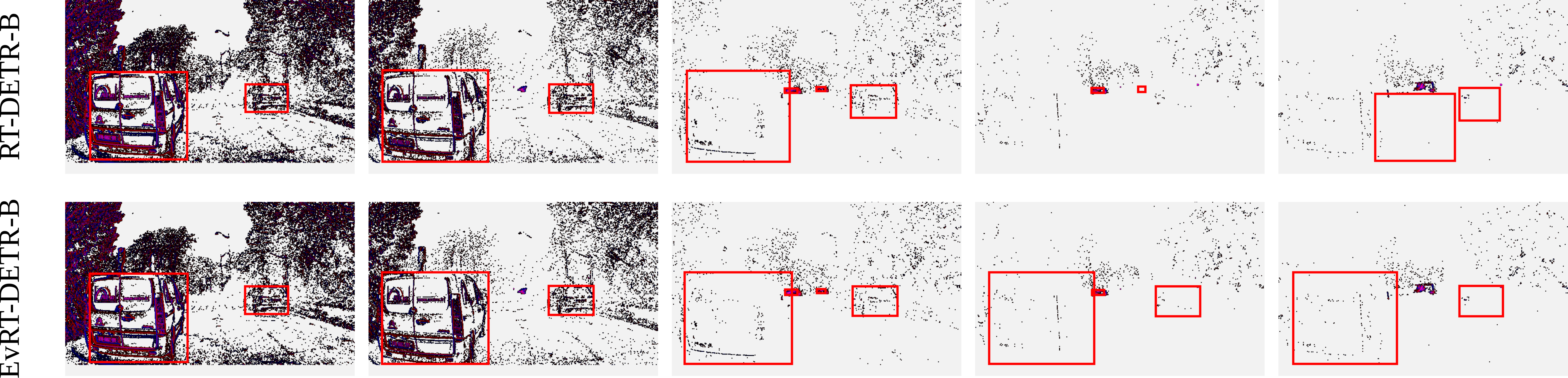}
    \caption{
        \textbf{Temporal Adaptation Benefits of EvRT-DETR.}
        Visualization of object detection across sequential \genFour frames captured by a vehicle-mounted camera that gradually comes to a stop. While the camera is in motion (left frames), both detectors maintain accurate bounding boxes of stationary vehicles. As camera motion decreases (middle frames), event data becomes increasingly sparse. When the camera stops completely (right frames), RT-DETR (top row) loses detection capability, while \thename (bottom row) maintains consistent detection by engaging its temporal memory module.
    }
    \label{fig:temporal_consistency}
\end{figure*}

\subsection{Temporal Modeling Considerations}

A key theoretical consideration for this adaptation framework is the choice of temporal memory mechanism. Event cameras present a unique challenge: static objects become ``invisible'' as they generate no events when motionless. Their detection requires having a persistent memory of past events. While transformers have dominated recent vision architectures, their fixed context windows limit temporal memory. We employ RNNs for their theoretically unbounded memory capacity through recurrent state, making them uniquely suited for maintaining object persistence during periods of invisibility in the event stream. This theoretical foundation has informed the \thename implementation.

\subsection{Technical Implementation of \thename}
\label{sec:method_technical}

The EvRT-DETR architecture implements the \tempname framework via strategic, minimal modifications to the RT-DETR model. As shown in \autoref{fig:evrt_detr}, we identify key integration points within the pre-trained model's latent space, where temporal information can be effectively incorporated without disrupting spatial feature extraction.

The RT-DETR encoder produces multiscale feature representations $\left\{\mathcal{E}_3, \mathcal{E}_4, \mathcal{E}_5 \right\}$, where each element corresponds to a transformer token encoding rich spatial information. These features, originating from the last three stages of the backbone $\left\{\mathcal{S}_3, \mathcal{S}_4, \mathcal{S}_5 \right\}$~\cite{zhao2024detrs}, provide an ideal latent manifold for temporal adaptation.

Following latent space adaptation principles, we introduce specialized temporal processing modules at each scale, implementing three parallel RNN-based memory modules $\{ \mathbf{R}_3, \mathbf{R}_4, \mathbf{R}_5 \}$.
This multiscale temporal adaptation design ensures that temporal dynamics are captured at different levels of feature abstraction. Critically, each RNN module interfaces with its corresponding encoder feature map through residual connections, mathematically expressed as:

\begin{align}
    \mathcal{E}_i^{t, proj} &= W_i^{down} \cdot \mathcal{E}_i^{t} \\
    \left( \mathcal{O}_i^{t, proj}, \mathcal{M}_i^t \right) &= \mathbf{R}_i \left( \mathcal{E}_i^{t, proj}, \mathcal{M}_i^{t-1} \right) \\
    \mathcal{O}_i^{t} &= W_i^{up} \cdot \mathcal{O}_i^{t, proj} \\
    \tilde{\mathcal{E}}_i^{t} &= \mathcal{E}_i^{t} + \alpha_i \cdot \mathcal{O}_i^{t},
\end{align}
where $W_i^{down} \in \mathbb{R}^{C_i^{RNN} \times C_i}$ and $W_i^{up} \in \mathbb{R}^{C_i \times C_i^{RNN}}$ are the projection matrices that transform features into and out of the temporal processing space, $\mathcal{M}_i^t \in \mathbb{R}^{C_i^{RNN} \times H_i \times W_i}$ represents the temporal memory state for the $i$-th feature scale at time $t$, $\mathcal{O}_i^{t}$ denotes the corresponding temporal adaptation output, and $\alpha_i$ is a learnable scaling factor that modulates the influence of temporal information on the spatial representation inspired by the ReZero technique~\cite{bachlechner2021rezero}. In our implementation, the input projection matrix $W_i^{down}$ is integrated within the RNN architecture, which already handles dimensionality reduction through its internal gating mechanisms. Meanwhile, the output projection $W_i^{up}$ is explicitly implemented to ensure proper integration back into the frozen RT-DETR architecture. Thus, the transformed feature representation $\tilde{\mathcal{E}}_i^{t}$ incorporates temporal context while maintaining the integrity of the original spatial representation $\mathcal{E}_i^{t}$. In addition, our ablation studies show that $C_i^{RNN}$ can be reduced 4x relative to $C_i$ with minimal performance impact.

For specific implementation of the RNN modules, we employ the ConvLSTM architecture~\cite{shi2015convolutional}, which effectively balances representational capacity with computational efficiency.
This choice was informed by ConvLSTM's demonstrated effectiveness in handling spatio-temporal data in the RVT model~\cite{gehrig2023recurrent}, allowing our adaptation to employ established temporal modeling techniques.

\section{Evaluation of the \thename Model}

\begin{table*}[!t]
    \centering
    \begin{tabular}{|c|c|c|c|c|c|}
\hline
\multirow{2}{*}{\textbf{Model}} & \multicolumn{2}{c|}{\textbf{Gen1}} & \multicolumn{2}{c|}{\textbf{1Mpx/Gen4}} & \multirow{2}{*}{\textbf{Params (M)}} \\
\cline{2-5}
  & mAP (\%) & Runtime (ms) & mAP (\%) & Runtime (ms) & {} \\
\hline
\rowcolor{gray!15}
AsyNet~\cite{messikommer2020event}           & 14.5 & - & - & - & 11.4 \\
\rowcolor{white} 
AEGNN~\cite{schaefer2022aegnn}               & 16.3 & - & - & - & 20.0 \\
\rowcolor{gray!15}
Spiking DenseNet~\cite{cordone2022object}    & 18.9 & - & - & - & 8.2 \\
\rowcolor{white} 
Events-RetinaNet~\cite{perot2020learning}    & 34.0 & - & 18.0 & - & 32.8 \\
\rowcolor{gray!15}
RED~\cite{perot2020learning}                 & 40.0 & 16.7 & 43.0 & 39.3 & 24.1 \\
\rowcolor{white} 
AEC~\cite{peng2023better}                    & 47.0 & 10.6 (3.9) & 48.4 & 37.6 (13.8) & 46.5 \\
\rowcolor{gray!15}
RVT-B~\cite{gehrig2023recurrent}             & 47.2 & 10.2 & 47.4 & 11.9 & 18.5 \\
\rowcolor{white} 
GET-T~\cite{peng2023get}                     & 47.9 & 16.8 & 48.4 & 18.2 & 21.9 \\
\rowcolor{gray!15}
S5-ViT-B~\cite{zubic2024state}               & 47.4 & 8.2 & 47.2 & 9.6  & 17.5 \\
\rowcolor{white} 
ASTMNet~\cite{li2022asynchronous}      & 46.7 & 35.6 & 48.3 & 72.3 & $>100$ \\
\rowcolor{gray!15}
SAST-CB~\cite{peng2024scene}           & 48.2 & - & 48.7 & 57.5 (19.7) & 18.9 \\
\rowcolor{white} 
ERGO-12~\cite{zubic2023chaos}          & 50.4 & 69.9 & 40.6 & 100.0 &  59.6 \\
\hline
\hline
\rowcolor{gray!15}
RT-DETR-T (ours)    & 46.0 & 6.4 & 44.1 & 8.6 & 20.1 \\
\rowcolor{white} 
RT-DETR-B (ours)    & 47.6 & 10.5 & 45.2 & 14.9 & 42.8 \\
\hline
\rowcolor{gray!15}
\thename-T (ours)   & \underline{52.3} & 8.4 & \underline{49.9} & 12.5 & 34.4 \\
\rowcolor{white} 
\thename-B (ours)   & $\mathbf{52.7}$ & 12.7 & $\mathbf{50.1}$ & 18.8 & 57.1 \\
\hline
    \end{tabular}
    \caption{\textbf{Performance Comparison on Event-based Camera Datasets.} Our approach achieves state-of-the-art results on both \genOne and \genFour benchmarks with significant improvements over previous methods. The first stage of the framework (RT-DETR-T/B) already achieves competitive performance using only image-based detection, while the proposed \tempname-adapted models (\thename-T/B) surpass all existing approaches. Runtime measurements are reported in milliseconds on an NVIDIA T4 GPU. Values in parentheses represent original reported times on different hardware with T4-equivalent times calculated using FLOPS/throughput conversion.}
    \label{tab:performance}
\end{table*}

The \thename model is evaluated on two standard EBC object detection benchmarks: \genOne~\cite{de2020large} and \genFour~\cite{perot2020learning} (also known as \mbox{Gen4}) datasets.
These datasets contain diverse driving scenarios captured by Prophesee's neuromorphic vision sensors, providing an ideal testbed for validating our adaptation approach under real-world conditions.

\subsection{Experimental Setting}

\paragraph{Datasets.} We evaluate on two standard Prophesee automotive datasets: \genOne~\cite{de2020large} and \genFour~\cite{perot2020learning}. \genOne contains 39+ hours of recordings at $304 \times 240$ resolution with manual annotations (1-4 Hz) for cars and pedestrians, presenting challenges with sparse labels and stationary objects. \genFour offers 14+ hours at higher resolution ($1280 \times 720$) with denser automatic annotations (60 Hz) for cars, pedestrians, and two-wheelers. Both datasets are organized as 60-second clips with standard train/test splits. \genFour's higher resolution and frame rate enable evaluation on more complex scenes, while \genOne's annotation pattern tests robustness to sparse temporal supervision.

\paragraph{Event Representations.} We adopt the well-established \textit{stacked 2D Histogram} approach from RVT protocols~\cite{gehrig2023recurrent}, constructing event frames from each $\SI{50} {ms}$ time window. Each frame is subdivided into 10 equally time-spaced $\SI{5} {ms}$ intervals, creating a representation with sufficient temporal resolution for our adaptation framework while remaining compatible with standard image processing architectures.

For the \genOne dataset (original size $(240, 304)$), we pad frames to $(256, 320)$ to ensure divisibility by 32 as required by the feature pyramid architecture of RT-DETR. For \genFour, we downsize from $(720, 1280)$ to $(360, 640)$ using bilinear interpolation and pad to $(384, 640)$. This maintains consistency with prior work while allowing for fair computational comparison.

\paragraph{Two-stage Training Process.} We train \thename in two stages: (1) training the RT-DETR detector on individual event frames using standard practices (Adam optimizer; EMA of weights) then (2) freezing these parameters and training only the lightweight RNN modules on both random and sequential clips to develop temporal memory. The first stage establishes robust spatial feature extraction, while the second adds temporal processing with minimal parameter overhead. For temporal training, we maintain RNN memory continuity across sequential clips while resetting for random clips using different clip lengths for \genOne (21 frames) and \genFour (10 frames). Full implementation details are provided in the supplementary material.

\paragraph{Evaluation Protocol.} We evaluate \thename using the standard COCO (Common Objects in Context) Mean Average Precision (mAP) measure~\cite{lin2014microsoft} that allows direct comparison with existing methods. For consistency with previous works, we use the EBC-specific implementation of the COCO metrics provided by Prophesee's Automotive Dataset Toolbox~\cite{psee_adt,perot2020learning}.
All inference times reported in Table 2 are standardized to NVIDIA T4 GPU performance for fair comparison.
Inference times include model forward pass and post-processing but exclude event preprocessing, following standard reporting practices. When prior work reported only preprocessing-inclusive times, we use their reported values directly.

\subsection{Performance Analysis}

\autoref{tab:performance} presents a comprehensive comparison of our \thename models against state-of-the-art approaches on the \genOne and \genFour benchmarks.

First, our baseline adaptation using two RT-DETR variants, RT-DETR-T (Tiny, with ResNet-18 backbone) and RT-DETR-B (Base, with ResNet-50 backbone), and trained on simple event frames achieves competitive performance ($46.0\%$ and $47.6\%$ mAP on \genOne; $44.1\%$ and $45.2\%$ mAP on \genFour) without requiring specialized architectures or complex event representations. This demonstrates that modern mainstream object detectors can be effectively adapted to event data through appropriate training.

More importantly, our complete \thename models with temporal adaptation significantly outperform all existing methods, achieving new state-of-the-art results on both benchmarks. \thename-B reaches $52.7\%$ mAP on \genOne ($+2.3\%$ over the previous best) and $50.1\%$ mAP on \genFour ($+1.4\%$ over the previous best). Even our lighter \thename-T model surpasses all existing approaches while maintaining faster inference times.

\autoref{fig:temporal_consistency} shows how the temporal adaptation module provides crucial capabilities for handling stationary objects -- a fundamental challenge for event cameras. When objects are in motion, both RT-DETR and \thename perform well, yet, when motion stops and event generation becomes minimal, \thename{}s temporal memory enables consistent detection.

These performance improvements come with minimal computational overhead. While we focus primarily on detection accuracy in this work, the runtime measurements in \autoref{tab:performance} demonstrate that our approach maintains competitive inference speeds despite adding temporal processing capabilities. 

\subsection{Framework Generalizability}

\begin{table}[!htb]
    \centering
    \resizebox{0.8\columnwidth}{!}{%
        \begin{tabular}{|l|cc|c|}
\hline
\multirow{2}{*}{Detector}  & \multicolumn{3}{c|}{mAP (\%)} \\
\cline{2-4}
{} & Original & I2EvDet & Improvement \\
\hline
\rowcolor{gray!15}
RT-DETR-T   & 46.0 & 52.3 & $+6.3$ \\
\rowcolor{white}
RT-DETR-B   & 47.6 & 52.7 & $+5.1$ \\
\hline
\rowcolor{gray!15}
YOLOX-T     & 36.0 & 42.4 & $+6.4$ \\
\rowcolor{white}
YOLOX-X     & 43.4 & 47.8 & $+4.4$ \\
\hline
        \end{tabular}
    }
    \caption{\textbf{Generalizability of \tempname Framework Across Different Object Detection Architectures on the \genOne Dataset}. Our temporal adaptation consistently improves performance regardless of the base detector architecture.}
    \label{tab:ablation_baseline}
\end{table}

To demonstrate that our results are not specific to the RT-DETR architecture, we apply the \tempname framework to YOLOX~\cite{ge2021yolox}. YOLOX is a CNN-based detector that contrasts with RT-DETR's transformer architecture and has been used in other EBC
applications~\cite{zubic2024state,gehrig2023recurrent}. \autoref{tab:ablation_baseline} shows that our temporal adaptation approach consistently improves performance across different baseline architectures on the \genOne dataset.
The substantial improvements (4.4-6.4 mAP) across both RT-DETR and YOLOX variants validate that \tempname represents a general framework for adapting image-based detectors to temporal domains rather than an architecture-specific optimization.
Additional details are provided in the supplementary material.

\subsection{Analysis of Adaptation Design Choices}
\label{sec:app_convlstm_ablation}

We systematically analyze key design choices in our temporal adaptation module to better understand the factors contributing to \thename's performance. This analysis provides insights into efficient adaptation strategies for event-based object detection. Additional ablation studies are provided in the supplementary material.

\begin{table}[!htb]
    \rowcolors{2}{white}{gray!15}
    \centering
    \resizebox{\columnwidth}{!}{%
        \begin{tabular}{|l|ccc|}
\hline
Model & mAP (\%) & $\text{mAP}_{50}$ (\%) & $\text{mAP}_{75}$ (\%) \\
\hline
RT-DETR-B (ours)          & 47.6 & 76.3 & 49.5 \\
\hline
\thename-B $( 0, 1, 1 )$   & $51.0$ & $80.7$ & $54.0$ \\
\thename-B $( 1, 0, 1 )$   & $52.2$ & $81.3$ & $55.1$ \\
\thename-B $( 1, 1, 0 )$   & \underline{$52.4$} & \underline{$81.7$} & \underline{$55.3$} \\
\hline
\thename-B $( 1, 1, 1 )$   & $\mathbf{52.7}$ & $\mathbf{82.0}$ & $\mathbf{56.0}$ \\
\hline
        \end{tabular}
    }
    \caption{\textbf{Effect of Temporal Module Placement.} Performance on the \genOne dataset with ConvLSTM modules at different feature scales. The array notation $(x,y,z)$ indicates presence (1) or absence (0) of temporal modules at scales $\mathcal{E}_3$, $\mathcal{E}_4$, and $\mathcal{E}_5$, respectively. Low-level features benefit most from temporal adaptation.}
    \label{tab:ablation_lstm_location}
\end{table}

\subsubsection{Multiscale Temporal Adaptation}

A crucial aspect of our approach is the insertion of temporal modules at multiple scales in the feature hierarchy. \autoref{tab:ablation_lstm_location} shows performance when selectively removing ConvLSTM cells from specific feature levels with the array notation $(x,y,z)$ indicating presence (1) or absence (0) at each scale $\{ \mathcal{E}_3, \mathcal{E}_4, \mathcal{E}_5 \}$.

The results reveal a clear pattern: the lowest-level features ($\mathcal{E}_3$) benefit most significantly from temporal adaptation as performance drops substantially when this module is removed. The impact diminishes at higher feature levels, suggesting that temporal information is most valuable for fine-grained, lower-level features that capture detailed motion patterns. This insight can guide more efficient adaptation designs, wherein computational resources are directed to where they provide maximum benefit.

\begin{table}[!htb]
    \rowcolors{2}{white}{gray!15}
    \centering
    \resizebox{\columnwidth}{!}{%
        \begin{tabular}{|l|ccc|c|}
\hline
Model & mAP (\%) & $\text{mAP}_{50}$ (\%) & $\text{mAP}_{75}$ (\%) & $N_\text{T}$ (M) \\
\hline
\thename-B (M=64)   & $52.1$ & $81.3$ & $54.9$ & $2.3$ \\
\thename-B (M=128)  & $52.5$ & $81.7$ & \underline{$55.6$} & $5.4$ \\
\thename-B (M=256)  & \underline{$52.7$} & $\mathbf{82.0}$ & $\mathbf{56.0}$ & $14.4$ \\
\thename-B (M=512)  & $\mathbf{52.9}$ & \underline{$81.9$} & $\mathbf{56.0}$ & $42.9$ \\
\hline
        \end{tabular}
    }
    \caption{\textbf{Impact of Temporal Module Capacity.} Performance on \genOne dataset with varying ConvLSTM hidden dimensions (M). The rightmost column shows the number of trainable parameters (in millions) in the temporal adaptation module.
    }
    \label{tab:ablation_lstm_memcap}
\end{table}

\subsubsection{Parameter Efficiency in Adaptation}

Inspired by parameter-efficient techniques such as LoRA (Low-Rank Adaptation)~\cite{hu2021lora}, we investigate the minimal temporal module capacity needed for effective adaptation. \autoref{tab:ablation_lstm_memcap} shows \thename-B performance with varying hidden dimensions in the ConvLSTM modules.

Remarkably, reducing hidden dimensions to 64 features (a 4x reduction from our baseline) results in only modest performance degradation ($-0.6\%$ mAP) while requiring just 2.3 M additional parameters -- merely $5.4\%$ of the base RT-DETR-B model's 42.8 M parameters. This configuration closely resembles LoRA-style adaptation, achieving substantial performance gains with minimal parameter overhead. Even our most parameter-efficient adaptation (M=64) outperforms all previous state-of-the-art approaches, demonstrating that effective temporal adaptation does not require extensive architectural modifications.
Increasing the hidden dimension to 512 features (42.9 M parameters, nearly doubling the model size) yields only a marginal $0.2\%$ mAP improvement, indicating diminishing returns.

\section{Conclusions}

This work introduces \tempname, a framework that efficiently transforms image-based detectors into video-capable models using minimal modifications. Applied to event-based cameras, our approach demonstrates that mainstream object detection architectures can be effectively adapted to temporal domains without specialized redesign. RT-DETR trained on simple event representations achieves performance comparable to specialized methods, while our \tempname-adapted model \thename achieves state-of-the-art results on \genOne ($+2.3$ mAP) and \genFour ($+1.4$ mAP) benchmarks. Ablation studies provide insights about optimal temporal module design and parameter efficiency. These results suggest that bridging mainstream computer vision and specialized domains can be achieved through targeted adaptation rather than complete architectural redesign, potentially accelerating progress across diverse temporal visual domains.

\section*{Acknowledgments}

The authors extend their sincere gratitude to Yuewei Lin (Brookhaven Lab) for his valuable feedback and suggestions. This work presented in this paper was funded by the U.S. Department of Energy (DOE), National Nuclear Security Administration, Office of Defense Nuclear Nonproliferation Research and Development. The manuscript has been authored by Brookhaven National Laboratory managed by Brookhaven Science Associates, LLC under the Contract No. DE-SC0012704 with the U.S. Department of Energy.

{
    \small
    \bibliographystyle{templates/ICCV2025-Author-Kit-Feb/ieeenat_fullname}
    \bibliography{refs}
}

\appendices
\section{Training Setup}

This section details the implementation of the \tempname framework as applied to event-based object detection. The training methodology reflects a two-stage adaptation approach: first establishing a robust spatial detector foundation then incorporating minimal temporal processing while preserving the pre-trained model's representation capabilities. This implementation demonstrates how mainstream detectors can be efficiently adapted to temporal data with minimal architectural changes

\subsection{RT-DETR}
\label{sec:app_train_rtdetr}

Our RT-DETR model is based on the reference PyTorch implementation\footnote{commit \texttt{5b628eaa0a2fc25bdafec7e6148d5296b144af85}}~\cite{rtdetr_repo,zhao2024detrs}.
We explore two RT-DETR configurations: RT-DETR-T (corresponds to RT-DETR-ResNet18) and RT-DETR-B (corresponds to RT-DETR-ResNet50).
The RT-DETR configurations match the reference implementations, except two modifications:
(1) we change the number of input channels of the backbones from 3 to 20, and
(2) we do not use ImageNet pre-trained backbones.
\autoref{tab:rtdetr_confs} summarizes the differences between the RT-DETR-ResNet18 and RT-DETR-ResNet50 configurations.

\begin{table}[!htb]
    \centering
    \resizebox{\columnwidth}{!}{%
        \begin{tabular}{|l|cc|}
\hline
Configuration & RT-DETR-T & RT-DETR-B \\
\hline
Backbone      & ResNet-18 & ResNet-50 \\
FPN Features  & $[128, 256, 512 ]$ & $[ 512, 1024, 2048 ]$ \\
Encoder Expansion & $\times 0.5$ & $\times 1.0$ \\
Decoder Layers    & 3 & 6 \\
\hline
        \end{tabular}
    }
    \caption{Comparison of the RT-DETR-T and RT-DETR-B Configurations.}
    \label{tab:rtdetr_confs}
\end{table}

We train all RT-DETR models for $400{,}000$ iterations using the Adam optimizer~\cite{kingma2014adam} with a batch size of 32 and a learning rate of $2 \times 10^{-4}$.
Similar to the reference RT-DETR training, we maintain an exponential moving average (EMA) of the model weights with a momentum of $0.9999$.
Unlike the reference RT-DETR training, we do not use any learning rate schedules or reduce the learning rate of the backbone relative to the encoder-decoder parts.

\paragraph{Environment.}
The RT-DETR training is conducted with \texttt{PyTorch-2.2.2} and \texttt{torchvision-0.17.2}
on a single NVIDIA RTX A6000 GPU.

\subsection{\thename}

The \thename-T and \thename-B models respectively extend RT-DETR-T and RT-DETR-B by adding ConvLSTM modules~\cite{shi2015convolutional}.
Our ConvLSTM blocks use a hidden dimension of 256 (matching the encoder feature maps) and a kernel size of 3. The temporal module outputs are integrated into the base model through residual connections with a scaling factor of 1.0 as we found learnable scaling parameters (similar to ReZero~\cite{bachlechner2021rezero}) provide no measurable performance benefit on the \genOne dataset.

The ConvLSTM modules are trained jointly for $200{,}000$ iterations with the Adam optimizer and a batch size of 8 while keeping the baseline RT-DETR models frozen.
Each batch contains 8 short clips of consecutive frames: 21 frames for \genOne dataset and 10 frames for \genFour dataset.
The number of frames per clip follows RVT's approach~\cite{gehrig2023recurrent}, except we increase the number of frames for the \genFour dataset from 5 to 10 for better performance.
Each batch contains 4 randomly sampled clips and 4 consecutive clips from videos.

Unlike the baseline RT-DETR training with a constant learning rate and EMA averaging, we do not apply EMA averaging to the ConvLSTM modules and use a learning rate scheduler instead.
For simplicity, we rely on the RVT-inspired one-cycle LR scheduler~\cite{smith2019super}.
Specifically, we use \texttt{PyTorch}'s \texttt{OneCycleLR} implementation of the scheduler, with the following parameters: maximum LR $2 \times 10^{-4}$, initial \texttt{div\_factor} 20, final \texttt{div\_factor} 500, \texttt{pct\_start} 0.0005, and annealing strategy ``linear.''

The ConvLSTM modules are trained in the same environment as the baseline RT-DETR networks using a single NVIDIA RTX A6000 GPU.

\subsection{YOLOX Baselines}

We implement YOLOX baselines to demonstrate the generalizability of our \tempname framework beyond RT-DETR.
We use the reference YOLOX backbone and head implementations\footnote{commit \texttt{ac58e0a5e68e57454b7b9ac822aced493b553c53}}~\cite{ge2021yolox,yolox_repo}.
The only architectural modification is changing the number of input channels of the backbone from 3 to 20.

To ensure fair comparison across architectures, we train YOLOX baselines using exactly the same setup as RT-DETR (cf.~\autoref{sec:app_train_rtdetr}). Specifically, we train for $400{,}000$ iterations using the Adam optimizer with a batch size of 32 and a learning rate of $2 \times 10^{-4}$. 

The inference is performed with NMS threshold of $0.65$ and confidence threshold of $0.01$.

\subsection{\tempname on YOLOX}

Following our \tempname framework, we extend the YOLOX baseline with temporal processing capabilities. We freeze the pre-trained YOLOX parameters and insert ConvLSTM modules between the PAFPN~\cite{liu2018path} neck and detection head.

The ConvLSTM modules follow the same configuration as \thename: hidden dimension of 256, kernel size of 3, and residual integration with scaling factor 1.0. Training follows the identical protocol as \thename with $200{,}000$ iterations using the Adam optimizer and OneCycleLR scheduler (max LR $2 \times 10^{-4}$). Each batch contains 4 random and 4 consecutive clips of 21 frames (\genOne) or 10 frames (\genFour).

\subsection{Notes on Two-Stage versus End-to-End Training}

While developing the \tempname framework, we experimented with both end-to-end and two-stage training approaches. For the end-to-end training, we combined the RT-DETR model with the ConvLSTM temporal adapters and trained them jointly. Our experiments indicate that the two-stage training approach enables significantly faster convergence compared to end-to-end training with much higher stability. Moreover, in our experiments, the two-stage training has no performance downsides compared to the end-to-end training.

Based on these observations, we adopt the two-stage training approach in this work. An additional benefit of two-stage training is it enables experimentation with different temporal modules on the same spatial detector baseline, accelerating development and providing clear separation between spatial and temporal performance components.

We note that our end-to-end training experiments did not involve extensive hyperparameter exploration, so it is possible that alternative hyperparameter configurations may improve end-to-end training performance. However, the two-stage approach has proven more robust and practical for our framework development.

\subsection{Data Augmentation Strategy}

While data augmentation is standard practice in computer vision~\cite{cubuk2018autoaugment,cubuk2020randaugment}, its application to event-based data has been limited~\cite{peng2023better}. Our adaptation framework treats EBC data as image-like frames, allowing us to leverage established augmentation techniques from mainstream computer vision.

Our augmentation strategy employs a standard chain of geometric transformations and random erasing as summarized in \autoref{tab:method_augs}. This approach bridges conventional image augmentation practices with the unique characteristics of event data, supporting the broader goal of adapting mainstream techniques to event-based vision.

\begin{table}[h!]
    \centering
    \rowcolors{2}{white}{gray!15}
    \resizebox{\columnwidth}{!}{%
    \begin{tabular}{|l|c|c|}
        \hline
        Augmentation & Magnitude & Probability \\
        \hline
        Random Horizontal Flip & -                  & 0.5 \\
        \hline
        Random Rotation        & $\pm 30^\circ$     & 0.6 \\
        Random Translation     & $\pm 0.5$          & 0.6 \\
        Random Scale           & $(0.5, 1.5)$       & 0.6 \\
        Random Shear           & $\pm 30^\circ$     & 0.6 \\
        \hline
        Random Erasure         & -                  & 0.4 \\
        \hline
    \end{tabular}
    }
    \caption{
    Standard augmentation chain adapted for event-based object detection. 
Each transformation is applied sequentially with the indicated probability to bridge mainstream vision techniques with event data processing.
}
    \label{tab:method_augs}
\end{table}

\autoref{tab:method_augs} shows our augmentation strategy, which leverages standard transformations from the \texttt{torchvision} package (\texttt{0.17.2})~\cite{torchvision2016}. Each augmentation is applied independently with its corresponding probability. When applied, its magnitude is randomly sampled from the specified range. For Random Erasure, we preserve the object labels while erasing image regions, maintaining detection supervision even in partially occluded scenarios.

\paragraph{Temporal Consistency in Augmentations.}

Our two-stage adaptation approach requires different augmentation strategies for each stage. For the base detector (RT-DETR) training on individual frames, we apply augmentations independently to each frame following standard practice. 

For the temporal adaptation stage, we carefully preserve temporal consistency by applying identical geometric transformations (flip, rotation, scale, translation, shear) across all frames in a video clip while still allowing per-frame variations through random erasing. This ensures that the temporal module learns meaningful motion patterns rather than artificially induced movements from inconsistent augmentations.

\section{Supplementary Ablation Studies}

This section provides additional ablation studies examining data augmentation strategies for event-based object detection and spatial context size in our temporal adaptation modules.

\subsection{Impact of Augmentation on Base Detector Performance}

When applying mainstream object detectors to event data, appropriate data augmentation becomes crucial for effective domain transfer. Our experiments show that without a proper augmentation strategy, the base detector performance is significantly compromised.

\begin{table}[!htb]
    \centering
    \rowcolors{2}{white}{gray!15}
    \resizebox{\columnwidth}{!}{%
        \begin{tabular}{|l|ccc|}
\hline
Model & mAP (\%) & $\text{mAP}_{50}$ (\%) & $\text{mAP}_{75}$ (\%) \\
\hline
RT-DETR-B (ours)         & $\mathbf{47.6}$ & $\mathbf{76.3}$ & \underline{49.5} \\
RT-DETR-B (no augs)      & 38.6 & 62.8 & 39.8 \\
\hline
RT-DETR-B (-Rotation)    & \underline{46.8} & 74.4 & 48.8 \\
RT-DETR-B (-Scale)       & 45.6 & 73.8 & 47.2 \\
RT-DETR-B (-Translation) & 45.0 & 72.6 & 46.2 \\
RT-DETR-B (-Shear)       & $\mathbf{47.6}$ & \underline{75.9} & $\mathbf{49.7}$ \\
RT-DETR-B (-Erase)       & \underline{46.8} & 75.1 & 48.7 \\
\hline
        \end{tabular}
    }
    \caption{Impact of data augmentation techniques on base detector performance for the \genOne dataset.}
    \label{tab:ablation_augs}
\end{table}

\autoref{tab:ablation_augs} quantifies the contribution of different augmentation techniques to our adaptation framework. Without augmentations, RT-DETR-B performance drops by 9 mAP points, highlighting their critical role in successful adaptation. Among individual transformations, spatial manipulations (translation and rescaling) provide the largest gains, suggesting that scale and position invariance are particularly important when adapting image detectors to event data. Random rotations and erasure techniques offer moderate improvements, while shear transformations show minimal impact.

\section{Spatial Context Size in Temporal Processing}

\begin{table}[!htb]
    \centering
    \rowcolors{2}{white}{gray!15}
    \resizebox{\columnwidth}{!}{%
        \begin{tabular}{|l|ccc|}
\hline
Model & mAP (\%) & $\text{mAP}_{50}$ (\%) & $\text{mAP}_{75}$ (\%) \\
\hline
\thename-B (KS=1)   & \underline{$52.3$} & \underline{$81.4$} & \underline{$55.4$} \\
\thename-B (KS=3)   & $\mathbf{52.7}$ & $\mathbf{82.0}$ & $\mathbf{56.0}$ \\
\thename-B (KS=5)   & $52.0$ & $81.3$ & $54.8$ \\
\hline
        \end{tabular}
    }
    \caption{Effect of ConvLSTM Kernel Size. Ablation study on \genOne dataset showing optimal temporal adaptation performance with $3 \times 3$ kernels.}
    \label{tab:ablation_lstm_ks}
\end{table}

Previous work on event-based object detection with ConvLSTM models, specifically RVT~\cite{gehrig2023recurrent}, found optimal performance with $1 \times 1$ ConvLSTM kernels (effectively pointwise LSTMs). However, our adaptation approach shows different optimal characteristics. As shown in \autoref{tab:ablation_lstm_ks}, \thename achieves best performance with $3 \times 3$ kernels, suggesting that spatial context is valuable when adapting frozen RT-DETR features. This demonstrates how adaptation design choices may differ from specialized architectures built from the ground up. Increasing to $5 \times 5$ kernels degrades performance, suggesting that the additional parameters and receptive field expansion do not provide beneficial information for the task given the available training data.

\section{Detailed YOLOX Results}

To demonstrate that our findings extend beyond transformer-based architectures, we evaluate our \tempname framework on YOLOX~\cite{ge2021yolox}, a CNN-based detector that has been used in other EBC applications~\cite{zubic2024state,gehrig2023recurrent}.

\begin{table}[!htb]
    \centering
    \rowcolors{2}{white}{gray!15}
    \resizebox{\columnwidth}{!}{%
        \begin{tabular}{|l|ccc|}
\hline
Model & mAP (\%) & $\text{mAP}_{50}$ (\%) & $\text{mAP}_{75}$ (\%) \\
\hline
RT-DETR-B            & $\mathbf{47.6}$ & $\mathbf{76.3}$ & $\mathbf{49.5}$ \\
\hline
YOLOX-T              & 36.0 & 59.9 & 36.3 \\
YOLOX-S              & 37.0 & 61.1 & 37.6 \\
YOLOX-L              & 43.1 & 69.6 & \underline{44.7} \\
YOLOX-X              & \underline{43.4} & \underline{69.7} & \underline{44.7} \\
\hline
\hline
\thename-B           & $\mathbf{52.7}$ & $\mathbf{82.0}$ & $\mathbf{56.0}$ \\
\hline
EvYOLOX-T            & 42.4 & 71.9 & 42.8 \\
EvYOLOX-S            & 43.6 & 73.1 & 44.4 \\
EvYOLOX-L            & 46.6 & 75.4 & 48.2 \\
EvYOLOX-X            & \underline{47.8} & \underline{75.7} & \underline{50.0} \\
\hline
        \end{tabular}
    }
    \caption{Performance comparison of RT-DETR and YOLOX variants with and without our \tempname temporal adaptation on the \genOne dataset. The consistent improvements across all architectures and model sizes demonstrate the generalizability of our adaptation framework beyond transformer-based detectors to CNN-based architectures.}
    \label{tab:ablation_baseline_detailed}
\end{table}

We apply our two-stage training approach to multiple YOLOX variants, using identical training configurations to our RT-DETR experiments. \autoref{tab:ablation_baseline_detailed} shows that YOLOX models achieve respectable baseline performance on the \genOne dataset with YOLOX-X reaching 43.4 mAP, which remains below RT-DETR-B's 47.6 mAP.

The \tempname framework provides substantial improvements across all YOLOX variants with gains ranging from 4.4 to 6.4 mAP. These consistent improvements demonstrate that temporal adaptation benefits extend across different architectural paradigms, validating our framework's generalizability. While transformer-based RT-DETR achieves higher absolute performance than CNN-based YOLOX variants, both architecture families benefit significantly from our temporal adaptation approach.

These results confirm that the \tempname framework represents a general strategy for adapting image-based detectors to temporal domains with broad applicability beyond transformer architectures. Testing with the most recent YOLO iterations could be an interesting direction for future work but is beyond the scope of this current study.

\begin{figure*}[t!]
    \centering
    \includegraphics[width=0.9\textwidth]{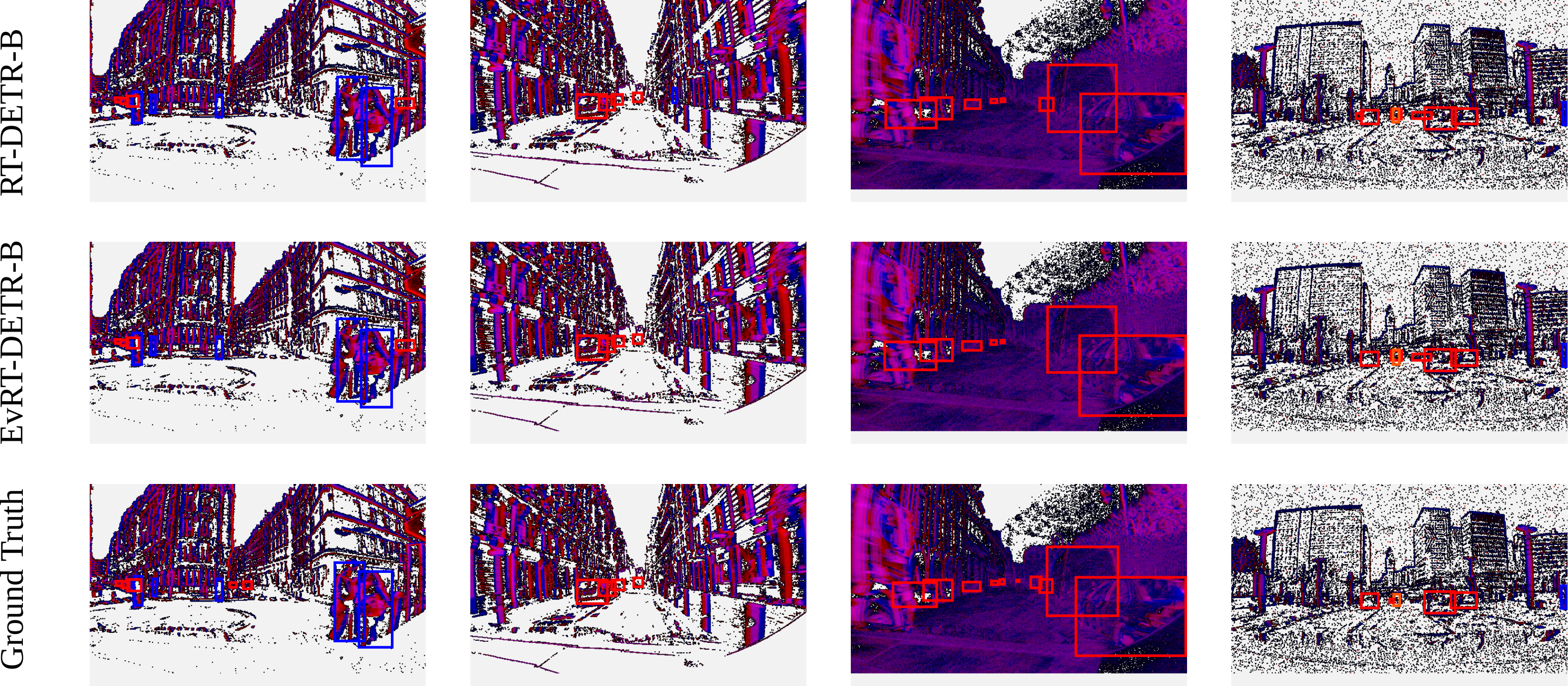}
    \caption{
        \textbf{Qualitative detection results on diverse \genFour automotive scenarios.} Top row: RT-DETR predictions. Middle row: EvRT-DETR predictions. Bottom row: Ground truth annotations. Bounding box colors indicate object classes: red (cars), blue (pedestrians), orange (two-wheelers). Both methods perform well on dynamic scenes with moving objects, demonstrating the effectiveness of our base RT-DETR adaptation to event-based data across varied lighting conditions and object configurations
    }
    \label{fig:random_plots}
\end{figure*}

\section{Adaptation Considerations for Higher Resolution Event Data (1Mpx)}

This section examines specific adaptation factors for the high-resolution \genFour dataset, focusing on resolution reduction techniques and temporal context length.

\subsection{Resolution Adaptation Strategy}

When adapting mainstream detectors to high-resolution event data, appropriate downsampling techniques become critical. For consistency with prior work, we reduce \genFour frames from $(720, 1280)$ to $(360, 640)$, but our investigation reveals that the interpolation method significantly impacts adaptation performance.

\begin{table}[!htb]
    \centering
    \rowcolors{2}{white}{gray!15}
    \resizebox{\columnwidth}{!}{%
        \begin{tabular}{|l|ccc|}
\hline
Model & mAP (\%) & $\text{mAP}_{50}$ (\%) & $\text{mAP}_{75}$ (\%) \\
\hline
RT-DETR-B (nearest)     & 42.3 & 71.8 & 42.3 \\
RT-DETR-B (bilinear)    & $\mathbf{45.2}$ & $\mathbf{75.1}$ & $\mathbf{46.0}$ \\
RT-DETR-B (bicubic)     & \underline{43.1} & \underline{72.2} & \underline{43.3} \\
\hline
        \end{tabular}
    }
    \caption{Impact of interpolation methods on base detector performance for the \genFour dataset downsampling.}
    \label{tab:ablation_interp}
\end{table}

\autoref{tab:ablation_interp} depicts how the nearest-neighbor interpolation substantially degrades RT-DETR performance, while bilinear interpolation yields optimal results. Interestingly, despite its theoretical advantages for natural images, bicubic interpolation proves less effective for event data. These findings align with observations from AEC~\cite{peng2023better} and highlight the importance of selecting appropriate domain transfer techniques when adapting mainstream vision models to event data.

\subsection{Temporal Context Length for Effective Adaptation}

The temporal dimension represents a critical aspect of our adaptation framework. While prior work like RVT~\cite{gehrig2023recurrent} uses 5-frame clips for temporal training, our experiments indicate that expanded temporal context benefits the adaptation process.

\begin{table}[!htb]
    \centering
    \rowcolors{2}{white}{gray!15}
    \resizebox{\columnwidth}{!}{%
        \begin{tabular}{|l|ccc|}
\hline
Model & mAP (\%) & $\text{mAP}_{50}$ (\%) & $\text{mAP}_{75}$ (\%) \\
\hline
\thename (5 frames)     & \underline{49.8} & \underline{80.7} & \underline{51.8} \\
\thename (10 frames)    & $\mathbf{50.1}$ & $\mathbf{80.9}$ & $\mathbf{52.1}$ \\
\hline
        \end{tabular}
    }
    \caption{Effect of temporal clip length on adaptation performance for the \genFour dataset.}
    \label{tab:ablation_gen4_clip_length}
\end{table}

\autoref{tab:ablation_gen4_clip_length} demonstrates the improvement achieved by extending the temporal window to 10 frames during the adaptation phase. This finding implies that providing longer temporal context during training allows the model to better capture persistent object representations across the temporal dimension, which is particularly important for event data where objects may be incompletely represented in shorter time windows.

\section{Comprehensive Evaluation Metrics}

To provide complete experimental validation, this section presents detailed performance metrics and additional visualizations of our models across both \genOne and \genFour datasets.

\subsection{Detailed COCO Metrics}

\begin{table*}[!t]
    \rowcolors{2}{white}{gray!15}
    \centering
    \begin{tabular}{|c|c|ccc|ccc|}
\hline
\textbf{Model} & \textbf{Data}
    & mAP ($\%$) & $\text{mAP}_{50}$ ($\%$) & $\text{mAP}_{75}$ ($\%$)
    & mAR ($\%$) & $\text{mAR}_{50}$ ($\%$) & $\text{mAR}_{75}$ ($\%$)
    \\
\hline
RT-DETR-T (ours)  & Gen1 & 46.0 & 74.4 & 47.2 & 38.3 & 53.2 & 38.3 \\
RT-DETR-B (ours)  & Gen1 & 47.5 & 76.3 & 49.5 & 39.9 & 54.6 & 42.1 \\
\thename-T (ours) & Gen1 & \underline{52.3} & \underline{81.4} & \underline{55.2}
    & \underline{44.2} & \underline{59.2} & \underline{51.8} \\
\thename-B (ours) & Gen1 & $\mathbf{52.7}$ & $\mathbf{82.0}$ & $\mathbf{56.0}$
    & $\mathbf{45.1}$ & $\mathbf{59.3}$ & $\mathbf{53.3}$ \\
\hline
\hline
RT-DETR-T (ours)  & 1Mpx & 44.1 & 74.2 & 44.3 & 33.5 & 50.7 & 52.6 \\
RT-DETR-B (ours)  & 1Mpx & 45.2 & 75.1 & 46.0 & 35.1 & 51.8 & 53.5 \\
\thename-T (ours) & 1Mpx & \underline{49.9} & $\mathbf{81.0}$ & \underline{51.6} 
    & \underline{38.6} & \underline{55.5} & $\mathbf{62.4}$ \\
\thename-B (ours) & 1Mpx & $\mathbf{50.1}$ & 80.9 & $\mathbf{52.1}$
    & $\mathbf{39.0}$ & $\mathbf{55.6}$ & 61.7 \\
\hline
    \end{tabular}
    \caption{\textbf{Complete COCO Evaluation Metrics.} Comprehensive performance comparison showing mAP and mAR metrics at IoU thresholds of 0.5:0.95 (default), 0.5, and 0.75 for all model variants on \genOne and \genFour datasets. Our temporal adaptation consistently improves both precision and recall across all thresholds and datasets, validating the robustness of the \tempname framework.}
    \label{tab:additional_performance}
\end{table*}

\autoref{tab:additional_performance} provides comprehensive COCO evaluation metrics including mean Average Precision (mAP) and mean Average Recall (mAR) at different Intersection over Union (IoU) thresholds. \tempname's temporal adaptation consistently improves performance across all metrics and datasets, demonstrating the robustness of our approach.

\subsection{Additional Visualizations}

\autoref{fig:random_plots} presents qualitative detection results on diverse \genFour automotive scenarios across varied lighting conditions and object configurations. Both RT-DETR and \thename demonstrate effective adaptation to event-based data representations, with comparable performance on dynamic scenes where objects generate sufficient event data through motion.

\begin{figure*}[t!]
    \centering
    \includegraphics[width=0.9\textwidth]{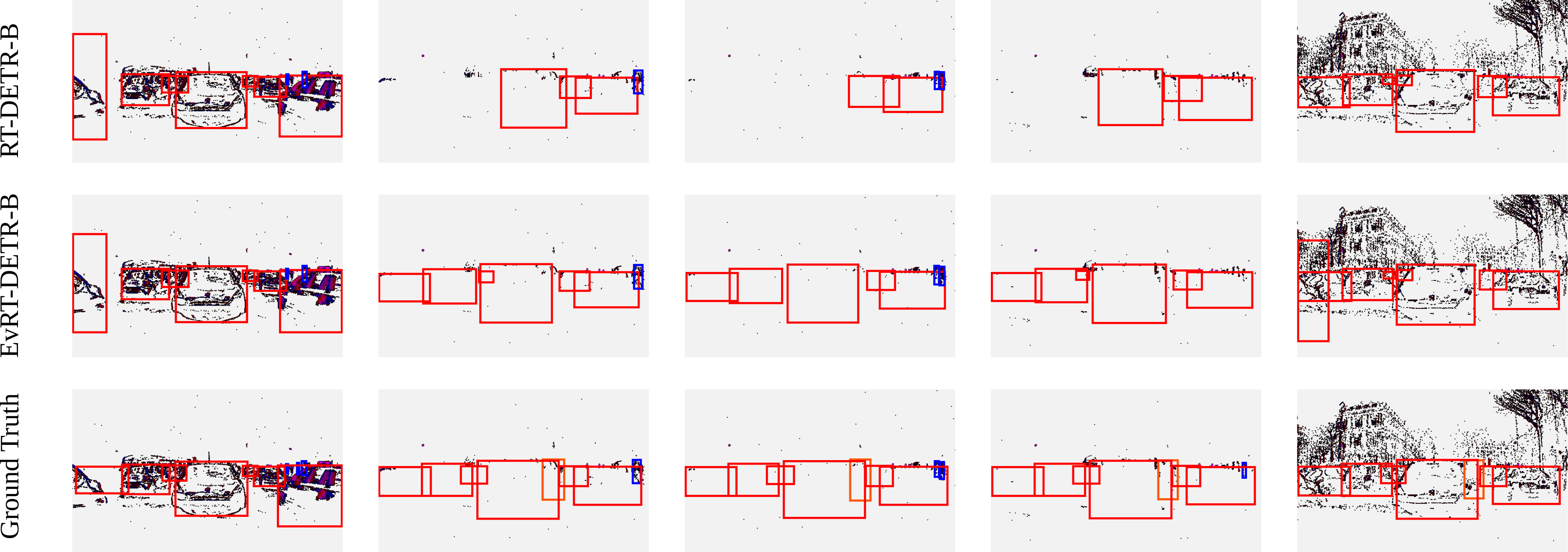}
    \caption{
        \textbf{Temporal sequence demonstrating event-based detection challenges during motion transitions}. A vehicle approaches an intersection, stops (creating sparse event data), then resumes motion. Frames shown every 100 frames for visualization clarity. Top row: RT-DETR predictions degrade significantly during stationary periods due to minimal event generation. Middle row: EvRT-DETR maintains more consistent detection by leveraging temporal memory from previous frames when objects were actively generating events. Bottom row: Ground truth annotations. Bounding box colors indicate object classes: red (cars), blue (pedestrians), orange (two-wheelers). While our temporal module substantially improves detection consistency during low-activity periods, challenges remain for heavily occluded objects, highlighting opportunities for future work
    }
    \label{fig:extra_temporal_consistency}
\end{figure*}

\autoref{fig:extra_temporal_consistency} illustrates the critical advantage of temporal memory during motion transitions. As vehicles stop at an intersection and event generation becomes sparse, frame-based detection degrades significantly while our temporal adaptation maintains consistent object localization by leveraging historical information. This sequence exemplifies the fundamental challenge that motivates our \tempname framework and demonstrates its effectiveness in real-world automotive scenarios.

\end{document}